# A Repeated Signal Difference for Recognising Patterns

Kieran Greer, Distributed Computing Systems, Belfast, UK.
http://distributedcomputingsystems.co.uk
Version 1.1

*Abstract -* This paper describes a new mechanism that might help with defining pattern sequences, by the fact that it can produce an upper bound on the ensemble value that can persistently oscillate with the actual values produced from each pattern. With every firing event, a node also receives an on/off feedback switch. If the node fires, then it sends a feedback result depending on the input signal strength. If the input signal is positive or larger, it can store an 'on' switch feedback for the next iteration. If the signal is negative or smaller, it can store an 'off' switch feedback for the next iteration. If the node does not fire, then it does not affect the current feedback situation and receives the switch command produced by the last active pattern event for the same neuron. The upper bound therefore also represents the largest or most enclosing pattern set and the lower value is for the actual set of firing patterns. If the pattern sequence repeats, it will oscillate between the two values, allowing them to be recognised and measured more easily, over time. Tests show that changing the sequence ordering produces different value sets, which can also be measured.

## 1   Introduction

This paper describes a new mechanism that could be used to more clearly define a pattern firing sequence, by the fact that it can produce an upper bound on the signal value that can persistently oscillate with the actual values produced from the pattern sequence. The mechanism is being written about because it might help to define signal differences in something that tries to model real neuronal activity. One question when simulating the human brain is how it is able to determine what is important, or distinguish between the different signals and activities. This was mentioned in [1] and is a well-known problem. It requires the pattern to be consistent and also distinct in its environment. The pattern changes are really the triggers for thinking or consciousness, so how would simple neurons be able to recognise them? The mechanism could also help to explain how binary signals can produce an analogue signal effect. This dichotomy might be attributed to some type of quantum process, but here it can be explained by more classical Newtonian mechanics. Memristors [2] are able to demonstrate a mechanism that shows how the most basic of organisms can store memories. The new mechanism of this paper also requires a memory component, where a neuron's firing ability is influenced by earlier firing events. It might also be of interest when modelling nested patterns, because the upper bound relates to the most neurons that would fire and also therefore to the largest pattern area. That area may then enclose smaller or nested patterns that also fire. The pattern sequence can repeat, as in a cycle, with a steady upper bound and a possibly increasing lower bound, making it easier for the simulator to tell what is important and what is not, because the signal differences are also repeated. Therefore, basic reinforcement of the difference would allow it to be recognised in some way. In this paper, the pattern also means the source of the input stimulus. The neurons can be modelled as typical neural network units and they require the same type of feedback signals, but work slightly differently to other neural network models. There are in fact 2 main differences for the model of this paper, as follows:

1. With every pattern firing event, every node receives an on/off switch feedback, for that pattern, as follows:
   a. If the node fires itself, then it sends the result of that as the feedback switch.
      i. If the signal fired is larger or positive, for example, it can store an 'on' switch for the pattern.
      ii. If the signal fired is smaller or negative, for example, it can store an 'off' switch for the pattern.
   b. If the node does not fire, then it does not affect the current feedback situation and receives the switch command produced by the last active pattern event for the same neuron.
2. The information that is fed back is only an on/off switch and does not contribute to the signal strength in any way. It works primarily at the input or synapse end of the neuron.

The rest of the paper describes these points in detail. The research of this paper has used an earlier counting mechanism [3] that provides a global and a local count. The global count is another upper bound value that would relate to every node firing every time. There is then the pattern sequence upper bound, as described in the following sections and the pattern sequence real value, so 3 measurements are possible now. The author still believes that the global upper bound is useful, but similar comparisons can be made without it. It is therefore probably not compulsory to use the counting mechanism to produce the test results and so it is up to the reader to decide exactly what the best mechanism might be.

The rest of this paper is organised as follows: section 2 describes some related work. Section 3 describes the new neuron structure and the firing event that it allows. Section 4 describes how the neuron might work as part of a network. Section 5 gives the results of some initial tests, while section 6 gives some conclusions on the work.

## 2   Related Work

This paper describes the new mechanism in terms of providing a stable way to maintain a notable difference in a signal, rather like a potential difference; so that a system trying to recognise pattern sequences, might be able to distinguish between them. It is also being put in the context of a real human brain. A lot of papers that address the problem [2][4 – 9] note factors such as oscillating signals, signal potential, signal stability and sensitivity to variance, and also conflicting excitatory and inhibitory signals providing a network balance. Some papers even note that noise is required for the neurons to fire properly. Noise can introduce a random value or change that might produce more graded signals, instead of all or nothing evaluations and also helps to contain the firing patterns [6]. The paper [2] describes a theory for how the Amoeba[1], the most simple of organisms, can learn and have a memory. The solution is to use a form of memristor and also to make use of biological oscillators. They explain that:

> 'The main idea behind functioning of this scheme is to use the internal state of memristor in order to store information about the past and control oscillations in the LC contour. In particular, we use the model of a voltage-controlled memristor, inspired by recent experiments, in which the resistance of memristor M can be changed between two limiting values $M_1$ and $M_2$, $M_1 < M_2$'.

The paper describes the process with differential equations and is not related to this work. Differential equations are not part of the author's current model, which has been mostly about describing clustering methods, perhaps changed by a non-mappable time element and mostly considering symbolic or conceptual data. A rate of change has not really been a factor, but it is part of other models, especially if they were to measure numerical instead of categorical values. Vogels et. al. [6] describe that theories have been created to explain how a network can maintain signal states, be balanced and still recognise the very small differences that would be the result of firing patterns. Weisbuch [7] is more cellular and describes the network properties in terms of automata with state change and logic gates. For these neural models, feedback is critical and Hopfield networks [10] in particular, have the type of feedback that this paper will suggest, or maybe the new Deep Learning networks [11].

---

[1] Also part of an invited talk by Prof. Wang at SAI'14, 'How will computers evolve over the next 10 years?'.

The paper [4] uses a Hopfield-modified network to perform what is known as factor analysis. It is possibly similar to cohesion or homogeneity, in that it takes the input signal vector and factors it into a low level signal space of relations or clusters. The low level factors would represent the first clustering stage. Their paper performs a Boolean factor analysis and they mention principle component analysis [12][13] and others, as alternatives. They also note that:

> 'During the learning stage, neurons that represent one common factor fire together when the factor appears in the signal. Due to the correlational Hebbian rule, these neurons become more tightly connected than those belonging to different factors. The different factor neurons are firing together only by chance (so, in the limit case, we can consider these factors as statistically independent). Thus, neurons of factors constitute attractors of network dynamics'.

The biological paper [8] discusses the role of dendritic spines. Pyramidal neurons[2] are looked at; where a neuron typically receives input from the dendrites and sends its output through the axon. The synapses are the connectors between these structures and have been used in earlier papers from the author to describe the connecting structures between neurons in general – shafts and connectors. Spines are then extensions of the dendrites with

---

[2] Santiago Ramón y Cajal: http://www.scholarpedia.org/article/Santiago_Ramón_y_Cajal.

a question of what they are for, if inputs can connect to the dendrites directly. The paper gives 3 main reasons for them and argues that these complement each other and are essential for a working distributed circuit. One reason for them is to filter synaptic potentials and electrically isolate input signals from each other. This would help to keep the signals distinct. It is also their purpose to add plasticity to the circuits of fixed neurons and connectors, where it might be difficult for a network to change after it has realised a large fixed structure. The paper [9] then extends the discussion and makes comparisons with the current artificial neural networks. It also notes the importance of an emergent function from ensembles, oscillations and other supporting features. One quote from that paper might be:

'As early as the 1930s, Cajal's disciple Rafael Lorente de Nó argued that the structural design of many parts of the nervous system is one of recurrent connectivity whose purpose could be to generate functional reverberations (patterns of neuronal activity that persist after the initial stimulus has ceased) among groups of neurons.'

## 3   Neuron Structure and Firing Event

The firing mechanism is based on ensembles of neurons that fire together. The neurons would typically be shared between patterns and sequences of input stimuli might therefore fire different sets of them. The scenario could therefore include nested patterns [14] that also share their neurons. The patterns determine the time-based input signals and the neurons can be influenced by earlier patterns, not just the current one. There is therefore probably the idea of a memory or capacitor, like a memristor might have. One pattern can be the largest and define the outer scope of the firing event. Implicit in this is nested patterns where, using the rules listed in the Introduction (section 1), smaller patterns overwrite their node subset only. It is equally allowed however to have a sequence moving from a smaller subset to a larger set again, but it would still all be part of the same ensemble for the mechanism that is described.

It can be assumed that only one pattern fires during each time unit, where Rule 1 of the Introduction states that if a neuron fires a stronger signal for a pattern, it sends feedback to ask it to fire again, the next time unit. If it fires a weaker signal, the request is not to fire next time. Stronger and weaker relates to a more variable type of on or off. As the mechanism is not final yet, there could be some flexibility with it. Rule 1 also states that if the neuron does not fire for the current pattern, it receives the switch command from the last pattern it did fire for. Note that the actual firing event can produce either zero or some input and so an earlier event can also force a weak feedback signal. Rule 2 states that this particular feedback is only the on/off switch command and does not affect the size of the output value. A schematic of the idea is illustrated in Figure 1, which relates to the same neuron shared between 2 pattern events, at times t1 and t2. The schematic is described further in section 3.1.

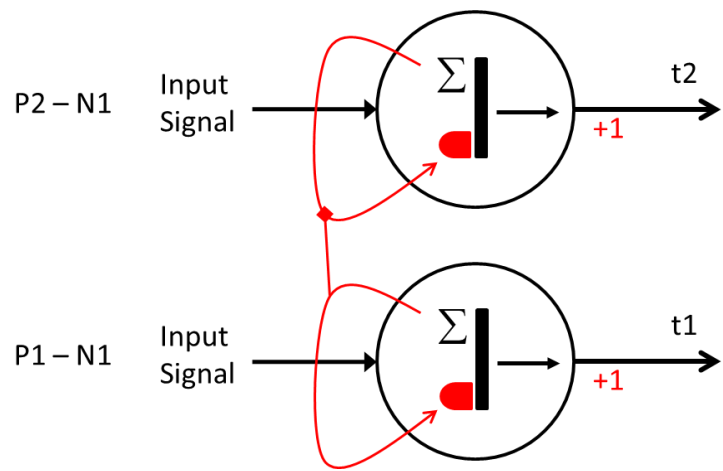

Figure 1. Schematic of the same neuron N1, shared between 2 different patterns P1 and P2. A single pattern fires during each time unit t1 and t2. The input signal from the pattern is summed and averaged for each event that the neuron fires on. If the neuron is active, it can store an on or off switch command for the next iteration. If the neuron feedback is not active, then the last active feedback switch is used. The counting mechanism could be used, with its global (pattern) and local (neuron) values.

In the diagram, the neuron is shown with a feedback loop from itself and also an imaginary connection to feedback from the earlier event. If you like, the neuron's feedback loop will reset the feedback from any earlier event. There is an input signal to the neuron and also an on/off switch (red block), managed by the feedback. The input is summed and may be activated by the switch, to produce a signal that activates the output. The output value itself is not influenced by the signal strength and could even be unit values, although it can be weighted. So if a neuron is active for a pattern, it uses its own output result as the feedback for the next time the pattern fires. This may be positive or negative. If a neuron is not active for a pattern, it can use an earlier pattern's value to set its switch, which would probably be positive. If looking at a cohesive set of neurons, the threshold trigger switches can be replaced by the switches of the neuron set that best cluster together.

### 3.1 Example

Consider the boolean case where the input stimuli are either 1 or 0 and the pattern neurons are all initially active. In this example, the value 0 actually means not to fire at all. The first stimulus pattern contains a value of 1 for input signal 1 and the second stimulus pattern contains a value of 0 for input signal 1. The first stimulus pattern executes (*t1*) and the feedback for signal 1 is positive, indicating to fire again next event. The second pattern executes (*t2*), but the 0 input value means that it does not trigger an output value that iteration, which also results in negative feedback. This is the end of phase 1 and for pattern 1 the neuron is still active but for pattern 2 it is now inactive. Pattern 1 then executes again (*t3*) and provides the positive feedback. Pattern 2 would then not fire the neuron (*t4*), but as it is inactive, the on switch command from pattern 1 tells it to fire. At the end of phase 2, the neuron is again active for both patterns. Pattern 1 then executes again (*t5*) and pattern 2 then activates the neuron (*t6*), but the weak input signal from pattern 2 means that it does not fire the output and the resulting negative feedback makes the neuron's switch inactive for pattern 2. During the fourth phase, pattern 2's inactive neuron uses pattern 1's switch value to tell it to fire again, and so on. Therefore, the total signal value increases to some maximum every other set of events and then back to a more true value for the patterns, every complementary set of events. If count updates are involved as some type of energy measurement, then this can increase slightly each time the maximum event is included, but it cannot increase beyond the maximum event value and is likely to level out before reaching that. The pattern sequence can cycle and there is a persistent difference between two value sets that can be measured.

## 4  Network Algorithms

The algorithms used in this paper are based on the equations listed in [15]. The intention was to test the theory of that paper further, but the new mechanism is worthy of mention, without considering all of that paper.

### 4.1  Main Algorithm

The main tasks are executed in the following order, where the following sections describe these tasks in more detail:

1. Retrieve the current cohesive set.
2. Update the weight values of all nodes in the cohesive set.
3. Determine a new cohesive set based on the new weight values.
4. Use the new cohesive set to update the counts.
5. After executing all patterns, retrieve the node values and calculate the energy value.

### 4.2  Cohesive Set for each Pattern

This is implemented using equations 1 to 5 in [15] and relates to section 4.1, task 3. Equation 5 of that paper simply clusters the nodes with the most similar reinforced values together. For these tests, only the top level of clustering was kept as the cohesive pattern. The idea would be that any other clusters might represent sub-clusters for that specific pattern, but the tests do not consider creating further hierarchies, for example. This is also a bit like principle component analysis. To avoid defining a 'cohesive difference' error amount, the weight reinforced values for each node can be sorted into descending order. Each node value can then be grouped with the one that it is closest to. This produces breaks in the node list, thereby resulting in more natural clusters. The first cluster that is created is then kept as the cohesive unit for the pattern.

#### 4.2.1  Cohesive Pattern Set Algorithm

In more formal terms, the process and can be defined as follows:

1. Before sorting any patterns, initialise the node-value cohesive set *CS* to be empty.
2. For each input pattern:
    a. Retrieve the current cohesive set for it.
    b. For each node in the cohesive set:
        i.  Retrieve the weight reinforcement value for the node.
        ii. Update the cohesive cluster *CS* to that value, for that node.
    c. Take the node-value cluster *CS* as the cohesive set for the current input pattern.

Where the following points can change the *CS* cluster values:

1. The re-ordering of the pattern node values will remove lower-valued weights. Therefore, if a node is included in the current cohesive set but has a smaller value, it will probably be removed from the new cohesive set.
2. If an input pattern does not include a node in its current set, but an earlier pattern does, then the earlier value remains as part of the current cohesive cluster *CS* and so it gets added for this input pattern as well.

### 4.3   Updating the Neuron Weight

This relates to section 4.1, task 2. Each node stores a reinforced weight value that is updated as follows:

1. If the node is part of the cohesive set for the related pattern:
    a. Retrieve the input value to the Node.
    b. Update or reinforce the weight using this input value. Note that the input can be 0, when it will not increase the weight value.

### 4.4   Updating the Counts

This relates to section 4.1, task 4. Each node also stores the counting mechanism, which is updated as follows:

1. For each pattern:
    a. Retrieve the set of cohesive nodes.
    b. If the node is in the cohesive set, update both the global and local counts.
    c. If the node is not in the cohesive set, update the global count only.

### 4.5   Calculating the Energy Value

This relates to section 4.1, task 5. The counting mechanism is used to determine an energy value for the input dataset or pattern firing sequence as a whole and also produces the oscillating values. This is calculated by producing averaged stats for the counts that have been generated, for all of the shared nodes. For these tests the global count is predictable and could even be a single value in each pattern. As there are 5 patterns with 5 nodes in each, it produces an average value of 5. The local count however is only updated if the node is part of the pattern's cohesive set, which can vary. The count values are all updated in unit increments of 1, with some local count values updated only every other iteration, for example. Then averaging over that will produce a smaller value.

## 5   Tests

A program has been written in Java to test some of the equations and algorithms described in section 4. It has been used to verify that the equations can at least work together, even if they are not part of a clustering program yet. Cohesion, or homogeneous node sets is the main test criterion, as a cohesive unit represents a cluster and that measurement is critical for producing the results of this paper. To update counts or weight values, binary data has been used, as described in Figure 2. There are 5 patterns in total, each with 5 values. Each binary value represents an input stimulus, for neuron 1 to neuron 5. The fact that pattern 1 and pattern 5 are the same is not important. The patterns are therefore the input values to a layer of 5 neurons that each store the weight reinforcement value and the counting mechanism. Each node has only one set values for all of the patterns.

If the input value is 1, then the neuron is likely to be part of the cohesive set and would update the weight and also the global and local counts each event. If the input value is 0, then the neuron is not likely to be part of the cohesive set. For this case, it would not update the weight value as it is 0. It does update the global count as that uses unit increments, but only updates the local count when an earlier trigger switch tells it to. Each update event is also counted, so that averaged totals can be produced.

*Input pattern 1*     1, 1, 0, 0, 1
*Input pattern 2*     1, 1, 0, 1, 0
*Input pattern 3*     1, 1, 0, 1, 1
*Input pattern 4*     1, 1, 0, 0, 0
*Input pattern 5*     1, 1, 0, 0, 1

Figure 2. Binary input stimulus dataset.

### 5.1   Counting Mechanism and Reinforcement Measurements

The tests still use the counting mechanism of [3] and for this paper, the cohesive measurement is actually the simplified Equation 5 of paper [15]. So cohesion is measured using a single reinforced value for each node, while the energy measurement is calculated from what the counting mechanism stores. Ignoring the global count value, there are then the two oscillating values that includes a second upper limit, relating to the maximum value for the actual set of pattern instances. This does not have to include every node, but relates to the largest or most enclosing sets of nodes. Then the other, lower oscillating value, is for each input pattern, as it actually is.

**5.2   Sustained Signal Difference**

This section describes measuring the count totals after each iteration, for the returned cohesive sets. The first cohesive set for each input pattern is the full set of nodes. The patterns are presented in order – from pattern 1 to pattern 5. After a full presentation and weight update process, the values for each node are retrieved and averaged, to give the first set of values, shown in Table 1, iteration 1. After the iteration, cohesive sets can be calculated for each pattern, because the reinforcement weights have some values. This leads to the maximum pattern value, shown in iteration 2. For iteration 3 then, some nodes with smaller input values have fired and been subsequently removed from the cohesive sets again. Iteration 2 however has updated the counts previously. This leads to a slight increase in the local count values that can occur maybe every other event. Iteration 4 is back to the full cohesive set again, iteration 5 the reduced or actual set again, and so on. So iterations 2, 4 and 6 are the upper bound and oscillating with that is iterations 1, 3 and 5 that increase slightly but may level out.

Table 1. The cohesive set is not cleared before being added to by the current pattern.

|  | Node 1 | Node 2 | Node 3 | Node 4 | Node 5 |
|---|---|---|---|---|---|
| **Iteration 1** | 5 | 5 | 0 | 2 | 3 |
| **Iteration 2** | 5 | 5 | 0 | 3 | 4 |
| **Iteration 3** | 5 | 5 | 0 | 2.666 | 3.666 |
| **Iteration 4** | 5 | 5 | 0 | 3 | 4 |
| **Iteration 5** | 5 | 5 | 0 | 2.8 | 3.8 |
| **Iteration 6** | 5 | 5 | 0 | 3 | 4 |

DCS                                                                                                  07 September 2016Table 2 shows what the values would be if the cohesive set is reset for every pattern. That is, there are no enclosing patterns. To show that pattern ordering is also important, Table 3 shows some values with the pattern ordering reversed. Node 4 has clearly been updated differently, even though the oscillating gap remains.

Table 2. The cohesive set is cleared before it is worked out for each pattern.

|             | Node 1 | Node 2 | Node 3 | Node 4 | Node 5 |
|-------------|--------|--------|--------|--------|--------|
| Iteration 1 | 5      | 5      | 0      | 2      | 3      |
| Iteration 2 | 5      | 5      | 0      | 2      | 3      |
| Iteration 3 | 5      | 5      | 0      | 2      | 3      |
| Iteration 4 | 5      | 5      | 0      | 2      | 3      |
| Iteration 5 | 5      | 5      | 0      | 2      | 3      |
| Iteration 6 | 5      | 5      | 0      | 2      | 3      |

Table 3. Same conditions as Table 1, except that the pattern order reversed. Node 4 produces a different set of values.

|             | Node 1 | Node 2 | Node 3 | Node 4 | Node 5 |
|-------------|--------|--------|--------|--------|--------|
| Iteration 1 | 5      | 5      | 0      | 2      | 3      |
| Iteration 2 | 5      | 5      | 0      | 2.5    | 4      |
| Iteration 3 | 5      | 5      | 0      | 2.333  | 3.666  |
| Iteration 4 | 5      | 5      | 0      | 2.5    | 4      |
| Iteration 5 | 5      | 5      | 0      | 2.4    | 3.8    |
| Iteration 6 | 5      | 5      | 0      | 2.5    | 4      |

## 6   Conclusions

This paper describes what appears to be a new method for neuronal firing sequences. It is being described because the result of it is unusual and might provide some effects that a neuronal network would like to have. The oscillating energy value can be sustained over the firing sequence, allowing for it to be measured more easily. This would make a simulation of the real brain more realistic. The justification for the mechanism has been put into the context of recognising pattern sequences. Other types of system that might find this mechanism interesting could be the cellular ones (the paper [16] might be relevant). The new type of neuron requires a feedback loop from the signal strength, but the feedback is an on/off switch, which does not contribute to the signal strength itself. Initial tests confirm that oscillating values are produced and could be compared to some type of electric potential difference, for example. It also fits in with the author's earlier work [1][14], as it would occur naturally in a scenario of nested patterns that fire in sequence, but also with a lot of shared neurons. The upper bound is a measure of the most enclosing patterns and the other value is for each pattern individually. The measurements are also sensitive to changes in the pattern ordering. While a computer program has been written to test the effect of the mechanism, it will probably be more of a long-term strategy, as part of the whole research cognitive model. It is not obviously useful by itself and would only be added to a model after a more basic framework was in place. But if there is an opportunity to model biological neurons more accurately, it will be considered.